\documentclass[letterpaper]{article} % DO NOT CHANGE THIS
\usepackage {aaai2026}  % DO NOT CHANGE THIS
\usepackage{times}  % DO NOT CHANGE THIS
\usepackage{helvet}  % DO NOT CHANGE THIS
\usepackage{courier}  % DO NOT CHANGE THIS
\usepackage[hyphens]{url}  % DO NOT CHANGE THIS
\usepackage{graphicx} % DO NOT CHANGE THIS
\usepackage{natbib}  % DO NOT CHANGE THIS AND DO NOT ADD ANY OPTIONS TO IT
\usepackage{multirow}
\usepackage{makecell}  % <====
\usepackage{caption} % DO NOT CHANGE THIS AND DO NOT ADD ANY OPTIONS TO IT
\usepackage{algorithm}
\usepackage{algpseudocode}
\usepackage{amsmath} 
\usepackage{amsfonts}
\title{Learning Depth from Past Selves: Self-Evolution Contrast for Robust Depth Estimation}
\author {
    Jing Cao\textsuperscript{\rm 1},
    Kui Jiang\thanks{Corresponding Author}\textsuperscript{\rm 1},
    Shenyi Li\textsuperscript{\rm 1},
    Xiaocheng Feng\textsuperscript{\rm 1},
    Yong Huang\textsuperscript{\rm 2}
}
\affiliations {
   \textsuperscript{\rm 1}Faculty of Computing, Harbin Institute of Technology,\\
   \textsuperscript{\rm 2}Key Lab of Smart Prevention and Mitigation of Civil Engineering Disasters, Harbin Institute of Technology\\
    \{caojing, 2022111995\}@stu.hit.edu.cn, \{jiangkui, huangyong\}@hit.edu.cn,
    xcfeng@ir.hit.edu.cn
}
\begin{document}

\maketitle

\begin{abstract}
Self-supervised depth estimation has gained significant attention in autonomous driving and robotics. However, existing methods exhibit substantial performance degradation under adverse weather conditions such as rain and fog, where reduced visibility critically impairs depth prediction. 
To address this issue, we propose a novel self-evolution contrastive learning framework called SEC-Depth for self-supervised robust depth estimation tasks.
Our approach leverages intermediate parameters generated during training to construct temporally evolving latency models. Using these, we design a self-evolution contrastive scheme to mitigate performance loss under challenging conditions.
Concretely, we first design a dynamic update strategy of latency models for the depth estimation task to capture optimization states across training stages. 
To effectively leverage latency models, we introduce a Self-Evolution Contrastive Loss (SECL) that treats outputs from historical latency models as negative samples. 
This mechanism adaptively adjusts learning objectives while implicitly sensing weather degradation severity, reducing the need for manual intervention. Experiments show that our method integrates seamlessly into diverse baseline models and significantly enhances robustness in zero-shot evaluations.
\end{abstract}


\section{Introduction}
Accurate depth estimation from images is a critical computer vision task with significant applications in 3D scene reconstruction \cite{yin2022towards} and autonomous driving \cite{zhong2022rainy,hong2025resilience}. However, the development of depth estimation techniques is hampered by the prohibitive cost of acquiring ground-truth depth annotations. To address this limitation, researchers have explored self-supervised approaches that recover depth cues from video sequences \cite{godard2017unsupervised,godard2019digging,zhou2017unsupervised} or stereo image pairs \cite{godard2017unsupervised,wang2023planedepth} using pose or photometric information (the consistency of pixel appearance under different viewpoints). 

Conventional self-supervised methods eliminate the need for annotations, but exhibit unreliable performance in adverse weather conditions. Weather particles violate photometric consistency assumptions, impeding real-world deployment. While some approaches investigate weather-invariant feature extraction \cite{vankadari2020unsupervised, liu2021self}, they lack generalization in diverse scenes. Subsequent knowledge distillation methods (transferring knowledge from a teacher model to a student model) \cite{gasperini2023robust,mao2024stealing,tosi2024diffusion} mitigate degradation by using pseudo-labels from clear-weather teachers. However, their independent training prevents effective knowledge transfer. 

\begin{figure}[t]
	\centering
	\includegraphics[width=1\linewidth]{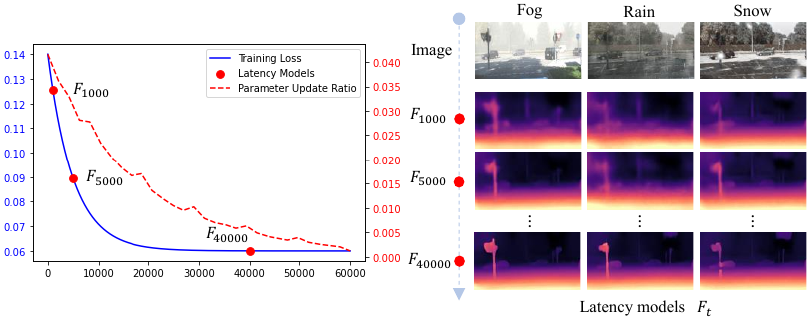}
    	\caption{Illustration of latency models evolution. The left figure shows the relationship between training step
        $t$, training loss, and parameter update ratio, where a decreasing update ratio indicates model convergence.  Models at different optimization steps $t$ within the parameter space are defined as latency models $F_t$. The right figure presents the depth outputs of latency models under adverse weather conditions. We leverage these evolving latency models to construct negative samples for the contrastive learning, which encourages the depth model to learn robust representations from its own historical information.}
	\label{fig:1}
    \vspace{-4mm}
\end{figure}

Contrastive learning (learning by enforcing depth consistency between clean and adverse conditions) presents a viable solution for robust depth estimation. Existing methods enforce depth consistency across scenes \cite{saunders2023self,wang2024weatherdepth} but risk collapsing solutions (degenerate output) by directly minimizing depth differences. 
D4RD \cite{wang2024digging} mitigates collapse by using diffusion sampling as anchors but requires extra distillation models and computational overhead. 
WeatherDepth \cite{wang2024weatherdepth} integrates contrastive learning %into a 
with progressive curriculum learning to balance scenes differences, but depends on presetdataset and complex curriculum schedule. 
To overcome these limitations, we propose an efficient plug-and-play contrastive framework that requires neither architectural modifications nor dataset priors. 

Inspired by the image restoration method \cite{wu2024learning}, 
we analyze the intrinsic optimization trajectory of self-supervised learning. 
Intermediate models in different training stages—termed latency models, exhibit progressive convergence toward optimal solutions (Figure~\ref{fig:1}). 
This historical state information provides model priors for contrastive learning. By incorporating these self-generated priors, we establish a self-evolution contrastive paradigm that avoids coupling with curriculum learning. 
Critically, latency models derive from the training process itself and are dataset-agnostic, ensuring generalization across diverse adverse conditions.
In addition, severe degradation causes erroneous depth predictions in specific regions, which complicates negative sample identification. To further address weather-induced local depth inconsistencies, we discretize depth maps into intervals and evaluate distributional similarity through probability consistency. This strategy better captures global depth characteristics than pixel-wise regression loss and improves distributional discrimination.

Building on the above analysis, we propose a self-evolution contrastive learning framework for self-supervised robust depth estimation (SEC-Depth). 
Our approach leverages latency models from historical parameters to generate self-evolution negative samples, while positive and anchor samples derive from the current model under clear and adverse conditions.  
We implement a dynamic latency model queue updated with recent parameters to maintain state-of-the-art historical representations.
Building on our interval-based depth consistency strategy, we propose a self-evolution contrastive loss that constructs anchor-positive-negative triplets from the model’s evolving trajectory and enhances robustness by contrasting current outputs with prior suboptimal predictions.

Unlike prior contrastive methods \cite{wang2024digging,wang2024weatherdepth}, our framework does not require baseline modifications and introduces generalization guidance through contrastive learning.
The main contributions of this work are:
\begin{itemize}
\item We propose a novel self-evolution contrastive framework for robust depth estimation across diverse conditions, compatible with existing self-supervised models without architectural changes.

\item We devise a dynamic latency model update strategy and a self-evolution contrastive loss, enhancing representation learning and stable learning in adverse scenes. 

\item Extensive experiments demonstrate significant improvements across multiple self-supervised tasks and strong zero-shot generalization (evaluation on unseen datasets) on six benchmarks.
\end{itemize}

% RELATED WORK
\section{Related work}
\subsection{Self-Supervised Depth Estimation}

Self-supervised depth estimation eliminates the reliance on depth annotations by leveraging geometric constraints from video sequences or stereo image pairs. Zhou et al. \cite{zhou2017unsupervised} pioneer this approach using geometric constraints between consecutive frames, establishing the foundation for monocular depth estimation. Subsequent research expands self-supervised learning using stereo pairs \cite{godard2017unsupervised,garg2016unsupervised} and videos \cite{godard2019digging,watson2021temporal}. Recent advances notably improve accuracy through data augmentation \cite{he2022ra,yao2024improving}, self-distillation \cite{marsal2024monoprob,wang2023planedepth}, multi-scale feature fusion \cite{liu2024self}, temporal fusion \cite{liu2024mono} and stronger network backbones \cite{zhang2023lite,zhao2022monovit}. 
Nevertheless, these methods often fail in complex weather conditions where factors such as illumination changes or precipitation violate the \textit{photometric consistency assumption}, limiting practical deployment.

\subsection{Robust Depth Estimation}

Achieving robustness of depth estimation is essential for real-world applications. Early approaches tackle domain adaptation via adversarial learning and image translation \cite{vankadari2020unsupervised,zhao2022unsupervised}, yet remain ineffective in adverse conditions (\emph{e.g.}, nighttime, rain). Subsequently, md4all \cite{gasperini2023robust} introduces knowledge distillation for robust training, while the latter works refine it through data augmentation \cite{tosi2025diffusion,mao2024stealing} and improves training strategies \cite{yan2025synthetic,jiang2025always}. However, student models remain fundamentally constrained by teacher performance.  
Recent methods employ contrastive learning (aligning similar representations while separating dissimilar ones) to enforce consistency between clean and degraded images. Robust-Depth \cite{saunders2023self} uses semi-augmented warping and bidirectional contrastive losses. WeatherDepth \cite{wang2024weatherdepth} applies curriculum learning (gradual exposure to harder samples) with progressive adaptation. D4RD \cite{wang2024digging} integrates multi-level contrastive learning with diffusion models.  
\textbf{Our work} distinguishes itself by proposing a plug-and-play contrastive framework that enhances robustness, which leverages latency models from historical parameters, combined with our interval-based depth consistency strategy. 
% METHOD
\begin{figure*}[!htbp]
    \hsize=\textwidth
    \centering
    \includegraphics[width=0.96\linewidth]{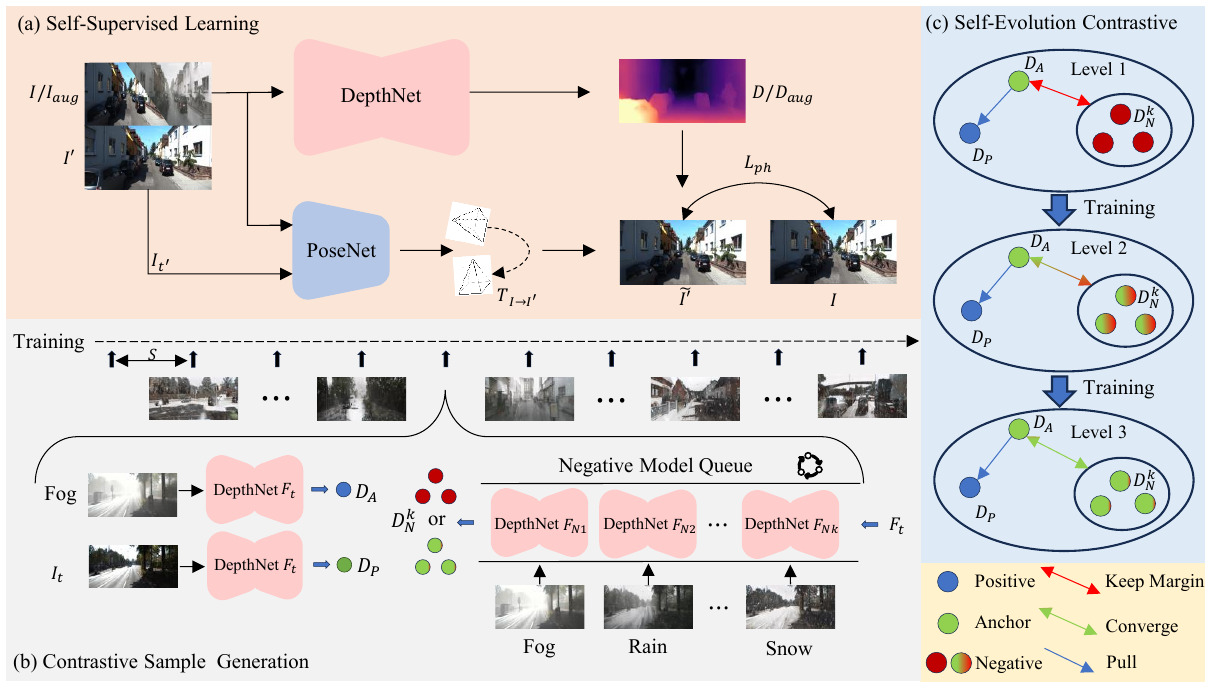}\vspace{-2mm}
%\vspace{-0.4cm}
    \caption{Illustration of our proposed pipeline. (a) Self-supervised learning is conducted on clean images. When augmented samples are introduced, the loss is computed using Equation (4).
    (b) During training, we maintain a model queue of size $j$, with parameters updated according to Algorithm 1.
    (c) As the self-supervised model continues to train, the parameters stored in the model queue gradually converge toward suboptimal states. Our self-evolution contrastive loss is designed to effectively leverage this parametric evolution.
    }
    \label{fig:2}
    \vspace{-4mm}
\end{figure*}
\section{Method}
\subsection{Preliminaries}
Self-supervised depth estimation methods predict a disparity map $D$ by leveraging geometric relationships between a target image $I$ and an auxiliary image $I'$. Using camera parameters, this disparity $D$ can be converted to a depth map $D'$. 
Denoting the depth estimation model as $F: I \rightarrow D \in \mathbb{R}^{W\times H}$, the network combines camera intrinsics $K$ and relative pose $T_{I \rightarrow I'}$ (obtained from either pose network or extrinsic parameters) to synthesize a warped image $\tilde{I'}$ from $I'$:
\begin{equation}
\tilde{I}' = I' \left\langle \text{proj}\left(D', T_{I \to I'}, K \right) \right\rangle,
\end{equation}
where $\left\langle \cdot \right\rangle$ denotes the pixel sampling operator. 
The depth prediction is constrained by a photometric reconstruction loss between $I$ and $\tilde{I'}$:
\begin{equation}
L_{ph} = \beta_1 \left(1 - \text{SSIM}(I,\tilde{I}')\right) + \beta_2 \left| I - \tilde{I}' \right|.
\end{equation}
Here, SSIM quantifies structural similarity between images. When processing augmented inputs $I_{\text{aug}}$, the warped image is computed using the corresponding depth ${D}_{\text{aug}}'$ following \cite{saunders2023self,wang2024weatherdepth}:
\begin{equation}
\tilde{I}' = I' \left\langle \text{proj}\left({D}_{\text{aug}}', T_{I \to I'}, K \right) \right\rangle. \label{eq:fol1}
\end{equation}

\subsection{Overview}
Our self-evolution contrastive learning framework operates independently from the self-supervised depth module. While the base model trains on clean scenes under normal conditions using Equation (1), we introduce a contrastive loss $L_c$ that directly leverages challenging weather conditions to enhance generalization. 

Given paired samples $(I, I_{\text{aug}})$ where $I$ is a clean image and $I_{\text{aug}}$ is its weather-corrupted counterpart (\emph{e.g.}, rain, snow or fog), both share identical scene content, but differ in appearance. The overall training objective is depicted as
\begin{equation}
L = L_{ph} + w L_c,
\end{equation}
where $w$ controls the contrastive weight. Crucially, $L_{ph}$ is calculated for both clean ($I$) and augmented ($I_{\text{aug}}$) images, while $L_c$ is applied \textit{only} when augmented images are processed. 
As shown in Figure~\ref{fig:2}, augmented samples $I_{\text{aug}}$ are periodically injected into training at fixed intervals $S$. This allows progressive adaptation to adverse conditions without disrupting the core self-supervised paradigm.

\subsection{Negative Samples Generation}
We maintain a queue of $j$ historical models (called \textit{latency models} or \textit{negative models}) $\{F_{N1}, F_{N2}, \dots, F_{Nj}\}$, initialized randomly before training. These are updated periodically using an exponential moving average (EMA) of the main model's parameters:
\begin{equation}
\theta_{k}^{*} = \omega \theta_{k}^{*} + (1 - \omega) \theta,
\end{equation}
where $\theta_{k}^{*}$ and $\theta$ are the parameters of the $k$-th negative and current main model, updated with a momentum of $\omega=0.01$.

To ensure that the negative model queue retains up-to-date information from the model during training, we adopt an adaptive update strategy for negative model queue.
The detailed update strategy for the negative model can be found in Algorithm 1.
The negative model queue is normally updated every $T$ steps. 
Additionally, when the model queue fails to generate sufficiently diverse negative samples (measured by the variance of their depth differences) for contrastive learning, we will proactively update it.
To generate negative samples, we randomly select $M$ (set to $j$ in our work) augmented images $I_{\text{aug}}$ and pass each through a randomly assigned negative model $F_{Nk}$, yielding disparity maps: 
\begin{equation}
D_N^k = F_{Nk}(I_{\text{aug}}), \quad k=1,2,\dots,j.
\end{equation}
These $\{D_N^k\}$ constitute the negative sample set.

Meanwhile, we define the disparity predictions of the self-supervised model on clean and augmented scenes as the anchor and positive sample, respectively:
\begin{equation}
D_A=F_t(I_{aug}), D_P=F_t(I),
\end{equation}
where $D_A$ and $D_P$ denote the disparity of the anchor and positive sample.
The $F_t$ represents the self-supervised model in training step $t$, and $I_{aug}$ is the augmented image (\emph{e.g.}, with rain, snow or fog).
Since augmented samples are calculated every $S$ steps, the model trains on clean images $I$ during standard self-supervised phases and switches to $I_{aug}$ when contrastive learning is applied.
In addition, only $D_A$ (anchor) retains gradients during contrastive loss computation.
We denote anchor, positive, and negative samples as $A$, $P$, and $N$ respectively in subsequent sections.

\subsection{Interval-Based Depth Distribution Constraint}

\begin{figure}[t]
	\centering
	\includegraphics[width=1\linewidth]{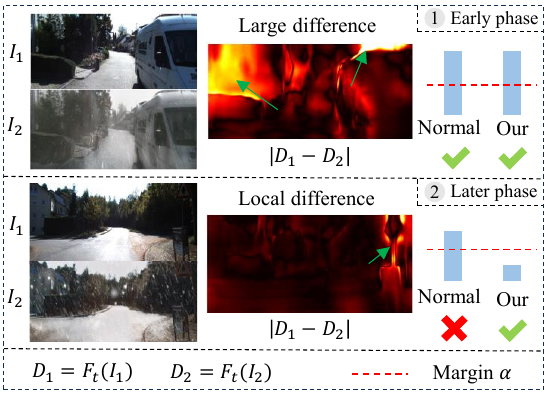}\vspace{-2mm}
	\caption{Illustration of the advantage of our interval-based depth modeling  strategy. (1) The model can reliably distinguish samples with significant overall depth differences. (2) Our strategy can better distinguish samples with local depth differences.}
	\label{fig:motivation}
    \vspace{-4mm}
\end{figure}

Existing contrastive learning methods for robust monocular depth estimation \cite{saunders2023self,wang2024weatherdepth} minimize depth discrepancies directly in the depth domain. However, this strategy fails to capture structural relationships when local distortions (\emph{e.g.}, object edges, or weather-induced degradations) dominate global depth distributions. Consequently, models struggle to assess relative depth distributions beyond pixel-wise errors, hindering effective use of negative samples. This limitation is illustrated in Figure~\ref{fig:motivation}, where variations in depth distribution impair the model’s relational judgment.

To address this issue, we propose a discretized depth modeling strategy that constructs domain-invariant distributional representations. %enhancing enhancing robustness across diverse domains.
Specifically, we divide the disparity range $[0, 1]$ into $N$ equal bins of width $1/N$, with bin centers $c_n = \frac{n + 0.5}{N}$ ($n = 0, 1, \dots, N-1$).
In this way, the continuous disparity values are converted into a discrete probability distribution. Furthermore, we use a Gaussian kernel ($\sigma = \frac{1}{2N}$) to assign each disparity value $d$ to bins, depicted as 
\begin{equation}
w_{n}(d)=\frac{1}{\sigma \sqrt{2 \pi}} \exp \left(-\frac{\left(d-c_{n}\right)^{2}}{2 \sigma^{2}}\right),
\end{equation}
where $\sigma=\frac{1}{2N}$ denotes the width of the bin. 
Subsequently, we generate the probability distributions by aggregating weights across all pixels and normalizing to obtain discrete distributions $P_X = \left[p_X^1, \dots, p_X^N\right]$ for anchor ($A$), positive ($P$) and negative ($N$) samples, where $\sum_{n=1}^N p_X^n = 1$.

\begin{algorithm}[htbp]
\caption{Negative Model Queue Update Strategy}
\begin{algorithmic}[1]
\Require Queue $Q$ of size $j$, current step $t$, update interval $T_{v} = 200$, anchor sample $a$, positive sample $p$, negative sample set $\mathcal{N}$, current model $\theta$
\State $n \gets 0$
\For{each training iteration}
    \State $\cdots$
    % \Statex \textit{Negative Model Queue Update}
    \If{$t \bmod T_v = 0$}
        \State \Call{UpdateQueue}{$Q, \theta$}
    \ElsIf{$ \mathbb{E}_{ n \sim \mathcal{N}}[\mathrm{Var}(D_{a} - D_{n}) ]  < \mathrm{Var}(D_{a} - D_{p})$}
        \State \Call{UpdateQueue}{$Q, \theta$}
    \EndIf
\EndFor

\Function{UpdateQueue}{$Q, \theta$}
    \State $Q[n] = \omega Q[n] + (1-\omega)\theta $
    \State $n \gets (n + 1) \bmod M$
\EndFunction
\end{algorithmic}
\end{algorithm}
\vspace{-3mm}

\subsection{Self-Evolution Contrastive Loss}
Our loss dynamically adjusts learning objectives using Jensen-Shannon (JS) divergence ($JS(\cdot||\cdot)$) to measure distributional similarity, formulated as:
\begin{equation}
L_{c}=JS\left(P_{A} \| P_{P}\right)+\frac{1}{M} \sum_{M}^{k} \left[\delta\Delta^{k}_1 +  JS\left(P_{A} \| P_{N}^{k}\right) \Delta^{k}_2 \right],
\end{equation}
\begin{equation}
\begin{aligned}
\Delta^{k}_i &= \max \left(\alpha_i - J S\left(P_{A} \| P_{N}^{k}\right), 0\right), i=1,2. \\
\end{aligned}
\end{equation}
The margin $\alpha_1$ and $\alpha_2$ controls the distance between the anchor and negative samples, allowing the model to adapt its optimization based on their discrepancy.
$\delta$ is the weight coefficient.

In our experiments, we observe that the self-supervised training process often converges rapidly to a suboptimal state during the early stages.
Therefore, we adopt a non-linear exponential decay strategy to adjust sample distances $\alpha_1$, allowing the model to better adapt to the evolving learning dynamics throughout training.
Then $\alpha_1$ is:
\begin{equation}
\alpha_1 = a e^{-15 \frac{t}{T}} + c,
\end{equation}
where $t$ is the current training step and $T$ is the total training step, $c$ and $a$ define the range of values of the parameter $\alpha_1$. We fix the value of $\alpha_2$ at 0.005 to assess whether the negative sample set reaches a convergent state.

To prevent destabilizing the self-supervised training process—particularly given the model's poor performance in both clear and degraded conditions during the early training stages, we initialize the contrastive loss weight with a small value ($w_s=0.01$) and gradually increase it as training progresses. The weight $w$ is defined as:
\begin{equation}
w = \begin{cases}
w_s (1+\max(0, e-e_a)) & e \leq e_b \\
w_s (e_b-e_a) & e > e_b,
\end{cases}
\end{equation}
where $e$ is the epoch number, $e_a=5$ and $e_b=15$.

\begin{figure*}[!htbp]
	\centering
	\includegraphics[width=0.99\linewidth]{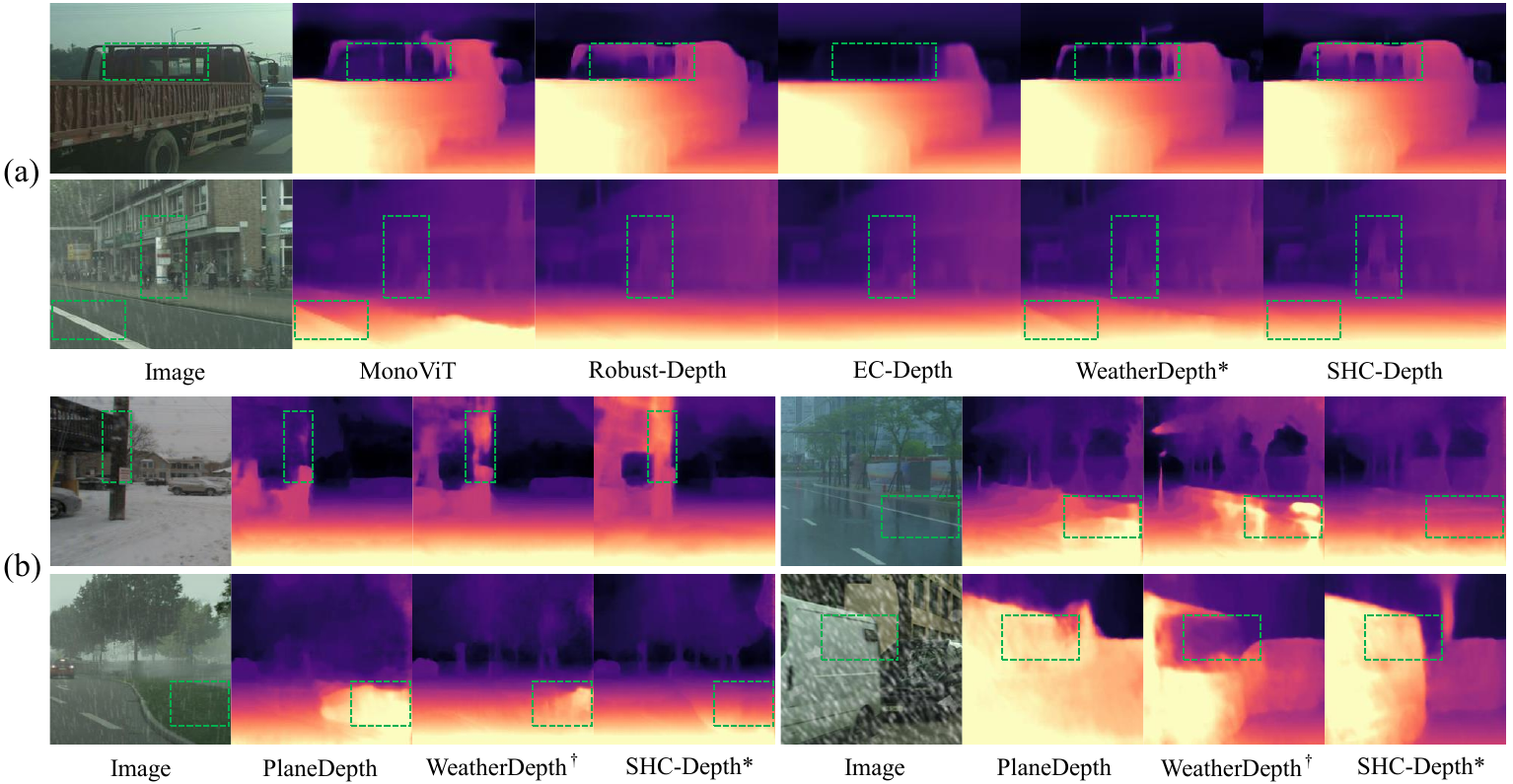}\vspace{-2mm}
	% \caption{Qualitative results for the zero-shot dataset. The first third and fourth rows correspond to fog, rain and snow scenes of Cityscapes dataset, and the second row is the fog scene of the DrivingStereo dataset. Qualitative results for other datasets can be found in the  supplementary materials.}
    \caption{(a) Qualitative results of DrivingStereo and Cityscapes dataset based on the MonoViT baseline. (b) Qualitative results of DrivingStereo and Cityscapes dataset based on the PlaneDepth baseline.}
    \vspace{-4mm}
	\label{fig:4}
\end{figure*}
% EXPERIMENTS
\section{Experiments}
\subsection{Datasets}
%\subsubsection{WeatherKITTI.}
\textbf{WeatherKITTI} \cite{wang2024weatherdepth} is a synthetic dataset derived from KITTI that includes six weather conditions: two rainy, two foggy, and two snowy scenes. For computing the contrastive loss, we select three representative conditions: mix\_rain/50mm, fog/75m, and mix\_snow/data. We follow Zhou's split \cite{zhou2017unsupervised}, which comprises 39810 training images and 4424 validation images. For testing, we use the Eigen split, consisting of 697 images.

\textbf{DrivingStereo} \cite{yang2019drivingstereo} is a real-world dataset. We evaluated the zero-shot performance of the model under challenging conditions using 500 test images each from foggy and rainy scenes.

\textbf{Cityscapes} \cite{cordts2016cityscapes} is a large-scale real-world dataset widely used in autonomous driving. We utilize three publicly synthetic datasets based on Cityscapes: Foggy Cityscapes \cite{sakaridis2018semantic}, SnowCityscapes \cite{zhang2021deep}, and RainCityscapes \cite{hu2019depth}.
Foggy Cityscapes contains 1,525 test images. Compared to the foggy scenes in DrivingStereo and WeatherKITTI, it exhibits more severe visibility degradation.
RainCityscapes is derived from the Cityscapes dataset and contains 262 training images and 33 testing images, all of which are normal weather images.
Then, three sets of parameters are used to simulate different degrees of rain and fog, attenuation coefficient (0.02,0.01,0.005), (0.01,0.005,0.01) and raindrop radius (0.03,0.015,0.002). 
In total, 1,188 test images are generated.
SnowCityscapes includes three snowfall intensities. From its original 2,000 test images, we select 1,510 samples with available ground-truth depth for zero-shot evaluation.

\textbf{Dense} \cite{bijelic2020seeing} is a real-world snowfall dataset.
Analysis in \cite{wang2024digging} demonstrated that the Dense is more suitable for depth estimation tasks than CADC \cite{pitropov2021canadian}.
We followed the processing method described in \cite{wang2024digging} and obtained 500 test images by sampling 1 in every 3 sequential images. After removing distorted areas, the final test images had a resolution of 1880$\times$924.

\subsection{Implement Details}
To validate the effectiveness of our self-evolution contrastive learning framework, we integrate it into two representative self-supervised depth estimation models: MonoViT \cite{zhao2022monovit} and PlaneDepth \cite{wang2023planedepth}. Experiments follow the original self-supervised training protocols for clear scenes (KITTI dataset). 
For the MonoViT baseline, we train for 30 epochs and test at 640$\times$192 resolution.
For PlaneDepth baseline, it is trained for 60 epochs using stereo inputs at 1280$\times$384 resolution (first-stage only).
Evaluation adheres to established test protocols: WeatherKITTI and DrivingStereo follow \cite{wang2024weatherdepth}; Foggy/Snow/Rain Cityscapes follow \cite{saunders2023self}; Dense follow \cite{wang2024digging}.
We set the hyperparameters $\delta=1e-4$, $a=0.05$, $c=0.001$, and the bin count $N=32$ (balancing runtime and accuracy). For contrastive learning, we use $j=3$ negative models.
Most other hyperparameters (batch size,learning rate and input image resolution, etc.) are the same as their baselines \cite{zhao2022monovit,wang2023planedepth}.

\begin{table}[!htb]\centering\scriptsize
%\label{monotest}
 \vspace{-0.2cm}
\scalebox{0.9}{
\begin{tabular}{c|ccccccc}
\hline
Method &  AbsRel &    SqRel &     RMSE & RMSElog &  $a_1$ &   $a_2$ &   $a_3$  \\
\hline
MonoViT &   0.120  &   0.899  &   5.111  &   0.200  &   0.857  &   0.953  &   0.980  \\
Robust-Depth &   0.107  &   0.791  &   4.604  &   0.183  &   0.883  &   0.963  &   \underline{0.983} \\
EC-Depth &   0.113  &   0.828  &   4.821  &   0.188  &   0.871  &   0.959  &   0.982 \\
EC-Depth* &   0.110  &   0.790  &   4.745  &   0.186  &   0.874  &   0.960  &   \underline{0.983} \\
WeatherDepth* &  \textbf{0.103} &  \textbf{0.738}  &  \textbf{4.414} &  \textbf{0.178}  &  \textbf{0.892} &  \textbf{0.965} &  \textbf{0.984}\\
SEC-Depth &   \underline{0.104}  &   \underline{0.762}  &   \underline{4.473}  &   \underline{0.180}  &   \underline{0.891}  &   \underline{0.964}  &   \underline{0.983} \\
\hline
\end{tabular}}
\vspace{-0.1cm}
\caption{\footnotesize Quantitative results on WeatherKITTI dataset using MonoViT as baseline. The best and second best results are marked in bold and underline.}
\label{WeatherKITTI}\vspace{-4mm}
\end{table}

\begin{table}[!ht]\centering\scriptsize
%\label{monozero}
\scalebox{0.88}
{
\begin{tabular}{c|ccccccc}
\hline
Method &  AbsRel &    SqRel &     RMSE & RMSElog &  $a_1$ &   $a_2$ &   $a_3$  \\
\hline
\multicolumn{8}{c} {\textbf{(a) DrivingStereo: Rain}}\\
\hline
MonoViT &   0.175  &   2.136  &   9.618  &   0.232  &   0.691  &   0.905  &   0.973  \\
Robust-Depth &   0.166  &   2.014  &   9.153  &   0.221  &   \underline{0.755}  &   0.939  &   0.982 \\
EC-Depth &   0.162  &   \textbf{1.723}  &   \textbf{8.478}  &   0.212  &   0.753  &   \textbf{0.948}  &   \textbf{0.986} \\
EC-Depth* &   0.162  &   \underline{1.746}  &   \underline{8.538}  &   0.212  &   \underline{0.755}  &   \underline{0.947}  &   \textbf{0.986}\\
WeatherDepth* &   \underline{0.158 } &   1.833  &   8.837  &   \underline{0.211}  &   {0.764} &   0.945  &  \underline{0.985} \\
\textbf{SEC-Depth} &  \textbf{0.157} &  1.820  &  8.999 &  \textbf{0.210}  &  \textbf{0.766} &  \textbf{0.948} &  \underline{0.985}\\
\hline
\multicolumn{8}{c} {\textbf{(b) DrivingStereo: Foggy}}\\
\hline
MonoViT& 0.109 & 1.204 & 7.760 & 0.167 & 0.870 & 0.967 & 0.990 \\
Robust-Depth& \textbf{0.105} & 1.132 & 7.273 &\textbf{ 0.150} & \textbf{0.882} & \textbf{0.974} &  \underline{0.992} \\
EC-Depth &   \underline{0.109}  &   {1.107}  &   \underline{7.230}  &   0.157  &   \textbf{0.882}  &  \textbf{ 0.974 } &   \underline{0.992} \\
EC-Depth* &   \textbf{0.105}  &   \textbf{1.061}  &  \textbf{7.121}  &   \underline{0.155}  &   \underline{0.880}  &   \textbf{0.974}  &   \textbf{0.994} \\
WeatherDepth* &   0.110  &   1.195  &   7.323  &   0.160  &   0.878  &  \underline{ 0.973}  &   \underline{0.992} \\
\textbf{SEC-Depth} & \textbf{0.105} & \underline{1.102} & 7.346 & 0.160 & \underline{0.880} & \underline{0.973} & \underline{0.992} \\
\hline
\multicolumn{8}{c} {\textbf{(c) Foggy Cityscapes}}\\
\hline
MonoViT& 0.156 & 1.877 & 9.598 & 0.245 & 0.770 & 0.910 & 0.967 \\
Robust-Depth& \underline{0.127} & \textbf{1.041} & \underline{6.617} & \underline{0.181} & \underline{0.846} & \underline{0.966} & \underline{0.991} \\
EC-Depth &   0.145  &   1.603  &   8.661  &   0.223  &   0.792  &   0.928  &   0.976 \\
EC-Depth* &   0.148  &   1.600  &   8.607  &   0.224  &   0.788  &   0.929  &   0.977 \\
WeatherDepth* &   0.131  &   1.214  &   7.089  &   0.189  &   0.833  &   0.959  &   0.989 \\
\textbf{SEC-Depth}& \textbf{0.122} & \underline{1.028} & \textbf{6.262} & \textbf{0.173} & \textbf{0.860} &\textbf{0.972} & \textbf{0.992} \\
\hline
\multicolumn{8}{c} {\textbf{(d) Rain Cityscapes}}\\
\hline
MonoViT& 0.181 & 0.403 & 2.158 & 0.277 & 0.711 & 0.907 & 0.959 \\
Robust-Depth & \textbf{0.151}  &   \underline{0.236}  &   \underline{1.589}  &   \underline{0.210}  &  \textbf{0.788}  &   0.954  &   \underline{0.987} \\
EC-Depth &   0.160  &   0.340  &   2.027  &   0.244  &   0.764  &   0.930  &   0.971 \\
EC-Depth* &   0.163  &   0.325  &   1.982  &   0.241  &   0.755  &   0.932  &   0.973 \\
WeatherDepth* & 0.155 & 0.253 & 1.708 & {0.214} & 0.773 & \underline{0.955} & 0.985\\
\textbf{SEC-Depth}  &\underline{0.154}  &   \textbf{0.222}  &   \textbf{1.473}  &   \textbf{0.203}  &   \underline{0.783}  &   \textbf{0.967}  &  \textbf{0.989} \\
\hline
\multicolumn{8}{c} {\textbf{(e) Snow Cityscapes}}\\
\hline
MonoViT& 0.229 & 2.686 & 10.512 & 0.319 & 0.599 & 0.851 & 0.942 \\
Robust-Depth &   0.158  &   1.757  &   8.504  &   0.231  &   0.766  &   0.928  &   0.978 \\
EC-Depth &   \underline{0.156}  &   \textbf{1.650}  &   8.302  &   \textbf{0.225}  &   \underline{0.773}  &   0.931  &   \textbf{0.980} \\
EC-Depth* &   0.158  &   \underline{1.651}  &   8.275  &   \textbf{0.225}  &   0.770  &   \underline{0.932}  &   \textbf{0.980} \\
WeatherDepth* & \textbf{0.155} & 1.672 & \underline{8.263} &\underline{0.226} & \textbf{0.779} & \textbf{0.933} & \underline{0.979}\\
\textbf{SEC-Depth}  &\underline{0.156}  &   1.712  &  \textbf{ 8.254}  &   0.227  &   \textbf{0.779}  &   \textbf{0.933}  &  \underline{0.979} \\
\hline
\multicolumn{8}{c} {\textbf{(f) Dense Snowy }}\\
\hline
MonoViT& 0.162 & 2.063 & \underline{9.797} & 0.262 & 0.762 & 0.904 &  {0.955} \\
Robust-Depth &   0.157  &   1.992  &   8.945  &   0.240  &   \textbf{0.786}  &   \underline{0.923}  &   \underline{0.971} \\
EC-Depth &   \textbf{0.154}  &   \textbf{1.866}  &   \textbf{8.801}  &   \textbf{0.236}  &   0.782 &   \underline{0.923}  &   \textbf{0.972} \\
EC-Depth* &   \underline{0.155}  &   \textbf{1.866}  &   \underline{8.828}  &   \underline{0.237}  &   0.780  &   0.922  &    \textbf{0.972} \\
WeatherDepth* & {0.157} & 2.000 & {9.021} & {0.243} &  \underline{0.781} & {0.919} & {0.968}\\
\textbf{SEC-Depth}  & 0.157  &   \underline{1.991}  &   8.856  &   {0.240}  &   \textbf{0.786}  &    \textbf{0.924}  &   \underline{0.971} \\
\hline
\end{tabular}
}
% \vspace{-2mm}
\caption{\footnotesize Zero-shot evaluation on DrivingStereo, Cityscapes and Dense datasets based on MonoViT baseline.}
\label{table:monozero} %\vspace{-2mm}
\end{table}

\subsection{Comparison Results Based on MonoViT}
In this section, we evaluate our method on the WeatherKITTI dataset and perform zero-shot testing on six additional datasets using methods based on the MonoViT baseline.
We compare the MonoViT baseline  \cite{zhao2022monovit}, as well as three MonoViT-based robust monocular depth estimation models: WeatherDepth* \cite{wang2024weatherdepth}, Robust-Depth \cite{saunders2023self} and EC-Depth \cite{song2023ec}. EC-Depth* is the second stage model in \cite{song2023ec}.
We use the pre-training parameters they have already released.

\subsubsection{WeatherKITTI Results.}

\begin{table}[!htb]\centering\scriptsize
 \vspace{-0.2cm}
\scalebox{0.9}{
\begin{tabular}{c|ccccccc}
\hline
Method &  AbsRel &    SqRel &     RMSE & RMSElog &  $a_1$ &   $a_2$ &   $a_3$  \\
\hline
PlaneDepth &   0.158  &   1.585  &   6.603  &   0.315  &   0.753  &   0.862  &   0.915  \\
WeatherDepth\textsuperscript{$\dagger$} &  \underline{0.099} &  \underline{0.673}  &  \underline{4.324} &  \textbf{0.185}  &  \textbf{0.884} &  \textbf{0.959} &  \textbf{0.981}\\
SEC-Depth* &   \textbf{0.098}  &  \textbf{0.652}  &  \textbf{ 4.392}  &   \underline{0.187}  &   \underline{0.883}  &  \textbf{0.959}  &   \textbf{0.981}  \\

\hline
\end{tabular}}
\vspace{-2mm}
\caption{\footnotesize Quantitative results on WeatherKITTI dataset using PlaneDepth as baseline.}
\label{table:stereotest}
\vspace{-2mm}
\end{table}

We show detailed comparative experiments on the WeatherKITTI datasets in Table~\ref{WeatherKITTI}.
Our framework demonstrates a 13.33\% decrease in AbsRel errors versus the MonoViT baseline. %compared to the baseline (MonoViT).
Compared with the robust alternatives (WeatherDepth* \cite{wang2024weatherdepth}, Robust-Depth \cite{saunders2023self}, EC-Depth \cite{song2023ec}), SEC-Depth achieved competitive results, which indicates the effectiveness of our method.

\begin{table*}[!htb] 
\renewcommand{\arraystretch}{1.2}
\centering
\footnotesize
\resizebox{.98\linewidth}{!}{
\begin{tabular}{cccc|ccccccc|ccccccc} %
\hline
\multicolumn{4}{c|}{Modules }  & \multicolumn{7}{c|}{WeatherKITTI} & \multicolumn{7}{c}{Zero-shot Datasets (Average)}\\
\hline
{CL}  & {ID} & $\Delta_1$& $\Delta_2$& AbsRel & SqRel &  RMSE & RMSElog & $a_1$ & $a_2$ & $a_3$ & AbsRel & SqRel&  RMSE & RMSElog & $a_1$ & $a_2$ & $a_3$  \\  
\hline
     &&&& 0.120 &0.899 &5.111 &0.200&0.857 &0.953 &0.980 &0.169 &1.728 &8.241 &0.250 &0.734 &0.907 &0.964         \\
    \checkmark&&&& 0.106& 0.835& 4.544& 0.183& 0.889 & 0.964 & 0.983 & 0.146 & {1.362} & 7.207 & 0.207& 0.802 &0.948 &0.983 \\
    \checkmark&\checkmark&&& 0.105& 0.796& 4.508& 0.181& 0.890 & 0.964 & 0.983 & 0.144 &1.340 & 6.913 & 0.204& 0.807 &0.950 &0.984 \\
    \checkmark& & \checkmark&\checkmark&  {0.105} & {0.807} &{4.523} &{0.181} &\textbf{0.891} &{0.964} &{0.983} &{0.144} &1.339 &{6.904} &{0.203}&{0.808} &{0.951} &{0.984}\\
    \checkmark&\checkmark&\checkmark&&  \textbf{0.104}& 0.772& 4.489& \textbf{0.180}& 0.890 & 0.964 & 0.983 & 0.143 & \textbf{1.309} & 6.893 & 0.203& 0.807 &0.951 &0.984     \\
\checkmark&\checkmark&\checkmark&\checkmark&  \textbf{0.104} & \textbf{0.762} &\textbf{4.473} &\textbf{0.180} &\textbf{0.891} &\textbf{0.964} &\textbf{0.983} &\textbf{0.142} &1.313 &\textbf{6.865} &\textbf{0.202}&\textbf{0.809} &\textbf{0.953} &\textbf{0.985}\\
\hline
% \bottomrule 
\end{tabular}}
%\vspace{-2mm}
\caption{Ablation study on individual components in SEC-Depth. %each module and stage mentioned in the pipeline. 
CL refers to the contrastive learning with degraded sample, ID refers to the Interval-Based Depth distribution constraint and $\Delta_i$ refers to our self-evolution contrastive loss.}
\label{table:ablation}
\vspace{-2mm}
\end{table*}

\begin{table}[!htb]\centering\scriptsize
\scalebox{0.88}
{   
\begin{tabular}{c|ccccccc}
\hline
Method &  AbsRel &    SqRel &     RMSE & RMSElog &  $a_1$ &   $a_2$ &   $a_3$  \\
\hline
\multicolumn{8}{c} {\textbf{(a) DrivingStereo: Rain}}\\
\hline
PlaneDepth &   0.215  &   3.659  &   12.112  &   0.271  &   0.670  &   0.889  &   0.964  \\
WeatherDepth\textsuperscript{$\dagger$} &   \underline{0.166}  &   \underline{1.874}  &   \textbf{8.844}  &   \underline{0.217 } &   \underline{0.748}  &   \underline{0.942}  &   \textbf{0.985}  \\
SEC-Depth* &  \textbf{0.157} &  \textbf{1.795}  &  \underline{8.976} &  \textbf{0.213}  &  \textbf{0.768} &  \textbf{0.944} &  \underline{0.984}\\
\hline
\multicolumn{8}{c} {\textbf{(b) DrivingStereo: Foggy}}\\
\hline
PlaneDepth &   \underline{0.122}  &   1.416  &   8.306  &   0.179  &   0.847  &   0.961  &   0.990  \\
WeatherDepth\textsuperscript{$\dagger$} &   0.123  &   \underline{1.404}  &   \underline{7.679}  &   \underline{0.172}  &  \underline{ 0.859}  &   \underline{0.968}  &   \underline{0.992}  \\
SEC-Depth* &  \textbf{0.110} &  \textbf{1.072}  &  \textbf{6.972} &  \textbf{0.158}  &  \textbf{0.876} &  \textbf{0.978} &  \textbf{0.994}\\
\hline
\multicolumn{8}{c} {\textbf{(c) Foggy Cityscapes}}\\
\hline
PlaneDepth &   0.172  &   1.922  &   9.157  &   0.255  &   0.747  &   0.918  &   0.966  \\
WeatherDepth\textsuperscript{$\dagger$} &   \underline{0.135}  &   \underline{1.014}  &   \underline{6.306}  &   \underline{0.181}  &   \textbf{0.873}  &   \underline{0.971}  &   \underline{0.992}  \\
SEC-Depth* &  \textbf{0.125} &  \textbf{0.905}  &  \textbf{5.999} &  \textbf{0.172}  &  \underline{0.857} &  \textbf{0.974} &  \textbf{0.993}\\
\hline
\multicolumn{8}{c} {\textbf{(d) Rain Cityscapes}}\\
\hline
PlaneDepth &   0.233  &   0.448 &   2.017  &   0.297  &   0.607  &   0.882  &   0.957  \\
WeatherDepth\textsuperscript{$\dagger$} &   \textbf{0.151}  &   \underline{0.195}  &   \underline{1.402}  &   \underline{0.198}  &   \underline{0.798}  &   \underline{0.967}  &   \underline{0.989}  \\
SEC-Depth* &  \underline{0.153} &  \textbf{0.194}  &  \textbf{1.366} &  \textbf{0.196}  &  \textbf{0.800} &  \textbf{0.969} &  \textbf{0.990}\\
\hline
\multicolumn{8}{c} {\textbf{(e) Snow Cityscapes}}\\
\hline
PlaneDepth &   0.395  &   5.809  &   14.803  &   0.519  &   0.362  &   0.636  &   0.805  \\
WeatherDepth\textsuperscript{$\dagger$} &   \underline{0.174}  &  \underline{1.680}  &   \underline{8.250}  &   \underline{0.246} &   \underline{0.739}  &   \underline{0.924}  &   \underline{0.972}  \\
SEC-Depth* &  \textbf{0.156} &  \textbf{1.473}  &  \textbf{7.824} &  \textbf{0.225}  &  \textbf{0.777} &  \textbf{0.939} &  \textbf{0.979}\\
\hline
\multicolumn{8}{c} {\textbf{(f) Dense Snowy}}\\
\hline
PlaneDepth &   0.175  &   2.140  &   9.205  &   0.263  &   0.754  &   0.911  &   0.962  \\
WeatherDepth\textsuperscript{$\dagger$}  &   \textbf{0.165}  &   \underline{1.877}  &   \textbf{8.269}  &   \textbf{0.238}  &   \textbf{0.775}  &   \textbf{0.928}  &   \textbf{0.973}  \\
SEC-Depth* &  \underline{0.168} &  \textbf{1.861}  &  \underline{8.387} &  \underline{0.245}  &  \underline{0.768}&  \underline{0.926} &  \underline{0.971}\\
\hline
\end{tabular}
}
\vspace{-1mm}
\caption{\footnotesize Zero-shot evaluation on DrivingStereo, Cityscapes and Dense datasets based on PlaneDepth baseline.}
\label{table:stereozero}\vspace{-2mm}
\end{table}

% \vspace{-4mm}
\subsubsection{Zero-shot Results.}
To further demonstrate the robustness of our model, we perform a zero-shot evaluation on six out-of-distributed datasets. 
As shown in Table~\ref{table:monozero}, our model achieves significant improvements over the baseline MonoViT under three synthetic and three real-world adverse weather conditions.
Compared with existing robust depth estimation methods based on MonoViT, our approach achieves state-of-the-art performance on most datasets.
Crucially, it surpasses WeatherDepth* (a contrastive learning benchmark) on the majority of evaluation metrics in 5/6 datasets, underscoring the superiority of our strategy. 
Visualizations in Figure~\ref{fig:4}(a) demonstrate detailed depth estimation in rain/fog and robust generalization to heavy rainfall, directly linking our framework’s design to improved generalization capability.

\subsection{Comparison Results Based on PlaneDepth}
We evaluate our method in the WeatherKITTI dataset and perform zero-shot testing on six additional datasets using the PlaneDepth baseline \cite{wang2023planedepth}. For distinction, \textbf{SEC-Depth* denotes our implementation using the PlaneDepth baseline}. We compare against PlaneDepth and its robust variant WeatherDepth\textsuperscript{$\dagger$} \cite{wang2024weatherdepth}, utilizing their released pre-trained parameters.

\subsubsection{WeatherKITTI Results.}
We show comparisons of our method against the baseline PlaneDepth and the robust WeatherDepth\textsuperscript{$\dagger$} in Table~\ref{table:stereotest}. It is obvious that our SEC-Depth* reduces AbsRel by 37.97\% compared to PlaneDepth. It also outperforms WeatherDepth\textsuperscript{$\dagger$}, indicating advantages when extending our strategy to other depth estimation approaches.

\subsubsection{Zero-shot Results.}
To assess generalization to unseen adverse weather conditions, we evaluate on six real-world and synthetic datasets. 
As shown in Table~\ref{table:stereozero}, our method 
significantly improves over PlaneDepth across all datasets and exhibits stronger zero-shot capability than WeatherDepth\textsuperscript{$\dagger$}, confirming its robustness.
Figure~\ref{fig:4}(b) displays the qualitative results, showing that our method accurately predicts the depth of buses in snowy scenes and effectively distinguishes reflections on water surfaces.

\subsection{Ablation}
We conduct ablation studies on WeatherKITTI and the six zero-shot datasets to validate our framework design. Due to space constraints, we report only MonoViT-based experiments, following the implementation details section. More ablation and analysis of our self-evolution contrastive frame
work will be demonstrated in the supplementary materials.

\subsubsection{Ablation on Major Design Components.}
Table~\ref{table:ablation} evaluates the contribution of individual components.
We augment the baseline with adversarially perturbed samples and contrastive learning (CL), while enforcing direct pixel-level depth alignment between clean and augmented data, which yields significant improvements. 

Subsequently, we introduce the Interval-Based Depth distribution constraint (ID) strategy (rows 3 and 4), which replaces the pixel-level depth alignment with depth distribution constraints (quantifying depth probability within intervals). 
Our ID strategy improves the RMSE and $\delta_1$ metrics, demonstrating its superiority over direct alignment in terms of both depth domain alignment and negative sample relation judgment.
In addition, by introducing the self-evolution contrastive loss ($\Delta_1$ and $\Delta_2$) to incorporate model priors and strategically select \textit{negative samples}, it further boosts performance on zero-shot datasets.
Collectively, these components demonstrate robust generalization in both synthetic and real-world adverse conditions.

\subsubsection{Ablation on Negative Step Selection.}
Table~\ref{table:negstep} examines the impact of negative step $S$, which controls the sampling frequency of augmented data in loss computation.
Smaller $S$ values increase exposure to challenging augmented samples during training, enhancing robustness but extending training time. In contrast, larger $S$ values reduce augmented sample utilization, consequently degrading generalization to adverse conditions.
Taking into account both accuracy and computational efficiency, we select $S=5$ in experiments.

\begin{table}[!ht]\centering\scriptsize
% \scalebox{0.88}
% {
\resizebox{0.47\textwidth}{!}{
\begin{tabular}{c|c|c|c|c|c|c|c|c}
\hline
Method &  AbsRel &    SqRel &     RMSE & RMSElog &  $a_1$ &   $a_2$ &   $a_3$ & Time \\
\hline
\multicolumn{9}{c} {\textbf{(a) WeatherKITTI}}\\
\hline
S=1 & 0.105& 0.819& 4.521& 0.181& 0.892& 0.964& 0.983& 26h10m  \\
S=5 & {0.104} & {0.762} & {4.473} & {0.180} & {0.891} & {0.964} & {0.983} &  21h45m  \\
S=10 &   0.105 &   0.794 &   4.525  &   0.181  &   0.890  &   0.964  &   0.983  &   20h35m  \\
S=20 &   0.106 &   0.801 &   4.544  &   0.182  &   0.887  &   0.963  &   0.983  &   19h28m  \\
\hline
\multicolumn{9}{c} {\textbf{(b) Zero-shot Datasets (Average)}}\\
\hline
S=1 & {0.141} & {1.329} &{6.773} &{0.199} &{0.816} &{0.954} &{0.985} & 26h10m \\
S=5 &  {0.142} &1.313 &{6.865} &{0.202}&{0.809} &{0.953} &{0.985}  & 21h45m\\
S=10 &   0.144 &   1.336 &   6.930  &   0.204  &   0.806  &   0.951  &   0.984  &   20h35m  \\
S=20 &   0.147 &   1.392 &   7.145  &   0.209  &   0.799  &   0.947  &   0.983  &   19h28m   \\
\hline
\end{tabular}
}
\caption{\footnotesize Ablation study of negative step selection.}\vspace{-5mm}
\label{table:negstep}
\end{table}

\section{Conclusion}
In this paper, we propose SEC-Depth, a novel self-evolution contrastive learning framework for self-supervised depth estimation. 
Our method implements a dynamic update strategy for historical depth models (latency models capturing prior training states) and constructs triplet samples (anchor, positive, and negative examples) using these models. To resolve ambiguous sample relationships in adverse weather, we transform disparity maps into binned probability distributions (discretized depth intervals for robust distributional comparison). Finally, we design a self-evolution contrastive loss that dynamically adapts optimization targets by contrasting current predictions against divergent outputs from historical models. Extensive validation across multiple datasets and baseline architectures confirms its transferability and effectiveness in zero-shot adverse conditions.

\section{Acknowledgments}
This research was financially supported by the National Natural Science Foundation of China (62501189, U23B2009), the Natural Science Foundation of Heilongjiang Province of China for Excellent Youth Project (YQ2024F006).

\bibliography{aaai2026}   

\appendix

\twocolumn

\section{Other Ablation Experiments Result}
\subsection{Computational Cost Analysis}
\begin{figure}[!htbp]
\centering
\includegraphics[width=0.85\linewidth]{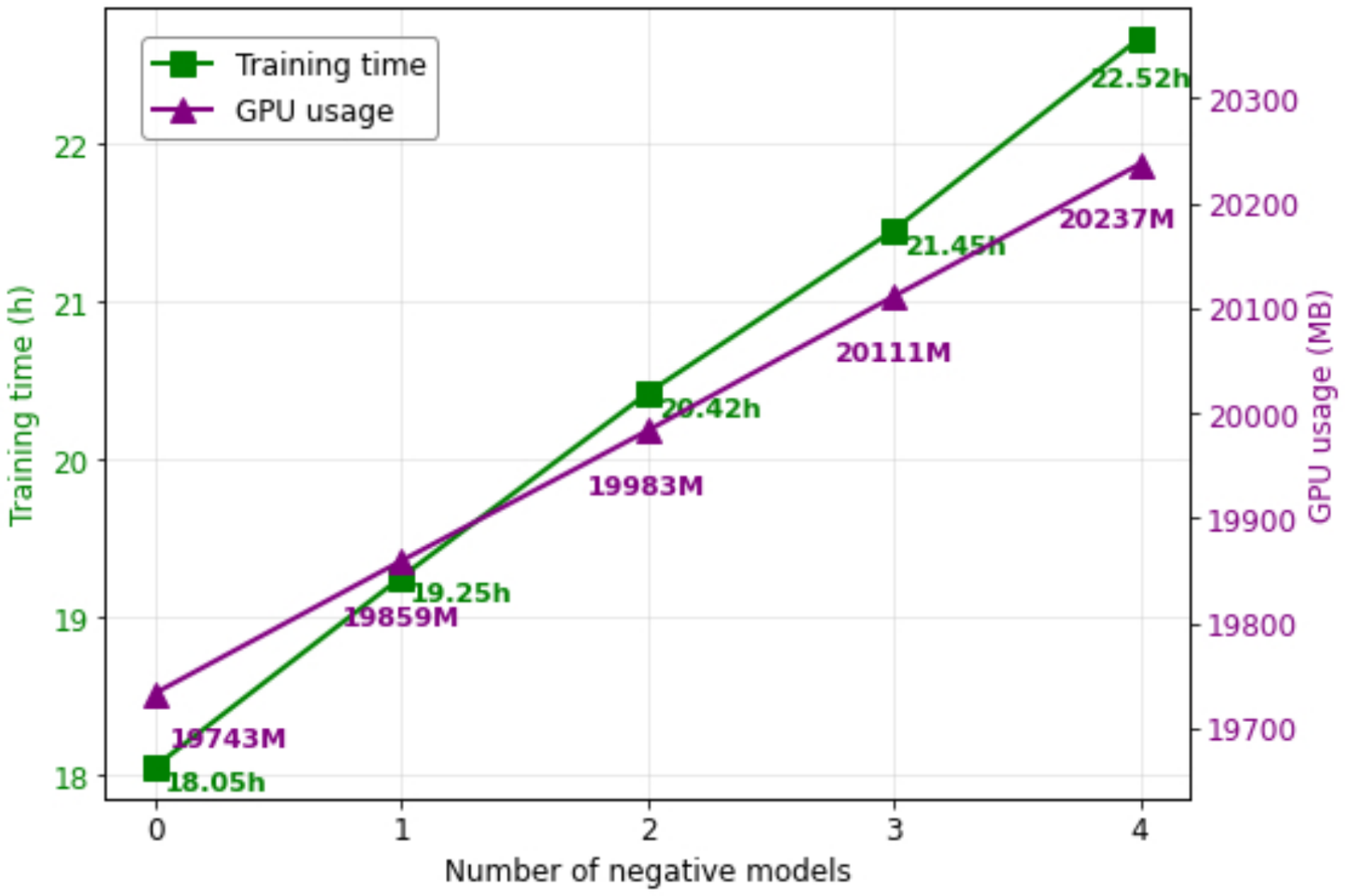}\vspace{-2mm}
\caption{The relationship between GPU memory usage, training time and the number of negative models}
\label{fig:cost}
\vspace{-3mm}
\end{figure}

We demonstrated the relationship between the number of negative models and GPU memory usage as well as training time on the MonoViT baseline, as shown in Figure~\ref{fig:cost}. 
All models are trained using a single NVIDIA RTX 3090 GPU.
Since negative models do not have gradient backpropagation, their GPU memory consumption primarily stems from storing model parameters, while the training time only includes the inference of negative samples and the computation of the loss function. 
The results demonstrate that when the number of negative models increases from 0 to 3, the training time increases by 18.84\% and the GPU memory usage increases by only 1.86\%.

\subsection{Ablation of $\alpha_1$}
In the self-evolution contrastive loss, we adopt a nonlinear exponential decay strategy to dynamically update margin $\alpha_1$ based on the convergence phenomenon observed during the self-supervised model training. 
Here, we replace the exponential decay strategy with a linear decay strategy, and the corresponding ablation results are shown in Table~\ref{table:alpha_1}.
The margin $\alpha_1$ is a variable to determine the relationship between the negative sample and anchor.
In our experiments, we observed a significant decline in performance when a linear decay strategy was applied.
We speculate that the model misjudges the relationships between samples, which interferes with the learning direction and leads to degraded performance.
Furthermore, we found that our method is highly sensitive to the accuracy of sample relationship judgments.
Incorrect relationships not only fail to enhance model performance but may also introduce adverse effects.
\begin{table}[!ht]\centering\scriptsize
\scalebox{0.88}
{
\begin{tabular}{c|c|c|c|c|c|c|c}
	\hline
	Method &  AbsRel &    SqRel &     RMSE & RMSElog &  $a_1$ &   $a_2$ &   $a_3$  \\
	\hline
	\multicolumn{8}{c} {\textbf{(a) WeatherKITTI}}\\
	\hline
	LD& {0.105}& {0.795}& 4.518& 0.181& {0.890}& 0.964& 0.983 \\
	ED (Ours)  & {0.104} & {0.762} &{4.473} &0.180 &{0.891} &{0.964} &{0.983} \\
	\hline
	\multicolumn{8}{c} {\textbf{(b) Zero-shot Datasets (Average)}}\\
	\hline
	LD & 0.147& 1.452& 7.241& 0.209& 0.799& 0.947& 0.983 \\
	ED (Ours) &  {0.142} &{1.313} &{6.865} &{0.202}&{0.809} &{0.953} &{0.985}\\
	\hline
\end{tabular}
}
\caption{\footnotesize Ablation study of $\alpha_1$. LD refer to the Linear Decay strategy for $\alpha_1$ and ED refer to the Exponential Decay strategy for $\alpha_1$}
\label{table:alpha_1}
\end{table}

\subsection{Ablation of $\alpha_2$}
We conducted an ablation study on the hyperparameter $\alpha_2$, which specifies the boundary that guides the convergence of the model.
The ablation results are presented in Table~\ref{table:alpha_2}.
When $\alpha_2$ is too large, the model may misunderstand the convergence of the negative model, thereby wrongly regarding negative samples as optimizable targets.
On the basis of our empirical observations, we use $\alpha_2=0.005$ as the default setting in all subsequent experiments, as it yields stable and robust performance.

\begin{table}[!ht]\centering\scriptsize
% \scalebox{0.88}
% {
\resizebox{0.47\textwidth}{!}{
	\begin{tabular}{c|c|c|c|c|c|c|c}
		\hline
		Method &  AbsRel &    SqRel &     RMSE & RMSElog &  $a_1$ &   $a_2$ &   $a_3$  \\
		\hline
		\multicolumn{8}{c} {\textbf{(a) WeatherKITTI}}\\
		\hline
		$\alpha_2=0$ & {0.104} & {0.772} &{4.489} &0.180 &{0.890} &{0.964} &{0.983} \\
		$\alpha_2=0.005$ & {0.104} & {0.762} &{4.473} &0.180 &{0.891} &{0.964} &{0.983} \\
		$\alpha_2=0.01$ &   0.105 &   0.787 &   4.501  &   0.180  &   0.891  &   0.964  &   0.983    \\
		$\alpha_2=0.02$ &   0.105 &   0.802 &   4.516  &   0.181  &   0.890  &   0.964  &   0.983    \\
		\hline
		\multicolumn{8}{c} {\textbf{(b) Zero-shot Datasets (Average)}}\\
		\hline
		$\alpha_2=0$ & {0.143} & {1.309} &{6.893} &0.203 &{0.807} &{0.951} &{0.984} \\
		$\alpha_2=0.005$ & {0.142} &{1.313} &{6.865} &{0.202}&{0.809} &{0.953} &{0.985}\\
		$\alpha_2=0.01$ &   0.143 &   1.357  &   6.901  &   0.203  &   0.808  &   0.952   &   0.985   \\
		$\alpha_2=0.02$ &   0.144&   1.379 &   6.947  &   0.204  &   0.807  &   0.951  &   0.984    \\
		\hline
	\end{tabular}
}
\caption{\footnotesize Ablation studies on margin $\alpha_2$ in loss function}
\label{table:alpha_2}
\end{table}

\subsection{Ablation of the bins count $N$}
We study the influence of the number of bins $N$, and results are presented in Table~\ref{table:bins}.
Selecting the number of bins requires balancing effectiveness against computational cost: more bins provide a finer partition but increase computational overhead.
Based on a global efficiency analysis, we found that 32 bins offer sufficient depth discrimination with low computational cost.
We therefore adopt 32 bins as the default parameter value in our experiments.
\begin{table}[!ht]\centering\scriptsize
% \scalebox{0.88}
% {
	\resizebox{0.47\textwidth}{!}{
		\begin{tabular}{c|c|c|c|c|c|c|c}
			\hline
			Method &  AbsRel &    SqRel &     RMSE & RMSElog &  $a_1$ &   $a_2$ &   $a_3$  \\
			\hline
			\multicolumn{8}{c} {\textbf{(a) WeatherKITTI}}\\
			\hline
			N=32 & {0.104} & {0.762} &{4.473} &0.180 &{0.891} &{0.964} &{0.983} \\
			N=64 & {0.105} & {0.786} & {4.494} & {0.180} & {0.891} & {0.964} & {0.983} \\
			N=128 &   0.104 &   0.758     &   4.508  &   0.180  &   0.890  &   0.964  &   0.983    \\
			\hline
			\multicolumn{8}{c} {\textbf{(b) Zero-shot Datasets (Average)}}\\
			\hline
			N=32  &  {0.142} &{1.313} &{6.865} &{0.202}&{0.809} &{0.953} &{0.985}\\
			N=64 &  {0.143} &1.327 &{6.883} &{0.202}&{0.809} &{0.952} &{0.985}\\
			N=128 &   0.141 &   1.320 &   6.852  &   0.202  &   0.808  &   0.953  &   0.985   \\
			\hline
		\end{tabular}
	}
	\caption{\footnotesize Ablation study of bins count $N$.}
	\label{table:bins}
\end{table}

\subsection{Ablation of negative loss parameters $\delta$}
In reference to Eq. (9), we examine the influence of varying negative loss weight coefficient $\delta$ and results are presented in Table~\ref{table:negpara}.
This coefficient controls the strength of the negative loss regularization term $\Delta_1$. When $\delta$ is too large (e.g., 1e-2), the regularization becomes dominant during the early stages of training and can hinder the model from converging to an optimal solution.
Based on experimental results, a value of 1e-4 achieves the best performance.
We used the same parameter configuration for all baseline models.

\begin{table}[!ht]\centering\scriptsize
% \scalebox{0.88}
% {
	\resizebox{0.47\textwidth}{!}{
		\begin{tabular}{c|c|c|c|c|c|c|c}
			\hline
			Method &  AbsRel &    SqRel &     RMSE & RMSElog &  $a_1$ &   $a_2$ &   $a_3$ \\
			\hline
			\multicolumn{8}{c} {\textbf{(a) WeatherKITTI}}\\
			\hline
			$\delta$=0 & {0.105} & {0.787} & {4.505} & {0.180} & {0.890} & {0.964} & {0.983} \\
			$\delta$=1e-2 & {0.106} & {0.803} & {4.530} & {0.182} & {0.889} & {0.964} & {0.983} \\
			$\delta$=1e-3 & {0.105} & {0.790} & {4.510} & {0.181} & {0.891} & {0.964} & {0.983} \\
			$\delta$=1e-4  & {0.104} & {0.762} &{4.473} &0.180 &{0.891} &{0.964} &{0.983} \\
			$\delta$=1e-5 &   0.104 &   0.757 &   4.477  &   0.180  &   0.891  &   0.964  &   0.983   \\
			\hline
			\multicolumn{8}{c} {\textbf{(b) Zero-shot Datasets (Average)}}\\
			\hline
			$\delta$=0 & {0.142} & {1.348} & {6.908} & {0.203} & {0.808} & {0.952} & {0.984} \\
			$\delta$=1e-2 & {0.143} & {1.327} & {6.923} & {0.204} & {0.807} & {0.951} & {0.984} \\
			$\delta$=1e-3 &  {0.142} &1.358 &{6.907} &{0.203}&{0.808} &{0.952} &{0.984}  \\
			$\delta$=1e-4  &  {0.142} &{1.313} &{6.865} &{0.202}&{0.809} &{0.953} &{0.985}\\
			$\delta$=1e-5 &   0.142 &   1.301 &   6.886  &   0.202  &   0.808  &   0.953  &   0.985 \\
			\hline
		\end{tabular}
	}
	\caption{\footnotesize Ablation study of negative loss parameters.}
	\label{table:negpara}
\end{table}

\subsection{Ablation of Data Augmentation}
Since our method does not rely on any preset dataset priors, we can freely control the severity of sample degradation.
To improve generalization, we randomly apply augmentations to degraded samples during training, including random erasing and random gaussian blur.
Meanwhile, no additional augmentations are applied to clean samples, ensuring consistency with the baseline training protocol.
We conducted an ablation study on the augmentation strategy, as shown in the Table ~\ref{table:augmentation}. Applying additional augmentations to degraded samples improves the model’s generalization ability.

\begin{table}[!ht]\centering\scriptsize
	\scalebox{0.88}
	{
		\begin{tabular}{c|c|c|c|c|c|c|c}
			\hline
			Method &  AbsRel &    SqRel &     RMSE & RMSElog &  $a_1$ &   $a_2$ &   $a_3$  \\
			\hline
			\multicolumn{8}{c} {\textbf{(a) WeatherKITTI}}\\
			\hline
			Ours w/o aug& {0.104}& {0.776}& 4.527& 0.181& {0.891}& 0.964& 0.983 \\
			Ours  & {0.104} & {0.762} &{4.473} &0.180 &{0.891} &{0.964} &{0.983} \\
			\hline
			\multicolumn{8}{c} {\textbf{(b) Zero-shot Datasets (Average)}}\\
			\hline
			Ours w/o aug& 0.143& 1.337& 6.981& 0.204& 0.806& 0.950& 0.984 \\
			Ours &  {0.142} &{1.313} &{6.865} &{0.202}&{0.809} &{0.953} &{0.985}\\
			\hline
		\end{tabular}
	}
	\caption{\footnotesize Ablation study of Data Augmentation.}
	\label{table:augmentation}
\end{table}

\end{document}